# Segmentation of the Prostatic Gland and the Intraprostatic Lesions on Multiparametic MRI Using Mask-RCNN

Zhenzhen Dai, Eric Carver, Chang Liu, Joon Lee, Aharon Feldman, Weiwei Zong, Milan Pantelic, Mohamed Elshaikh, Ning Wen

Department of Radiation Oncology, Henry Ford Health System, Detroit, MI, United State

Abstract

Prostate cancer (PCa) is the most common cancer in men in the United States. Multiparametic magnetic resonance imaging (mp-MRI) has been explored by many researchers to targeted prostate biopsies and radiation therapy. However, assessment on mp-MRI can be subjective, development of computer-aided diagnosis systems to automatically delineate the prostate gland and the intraprostatic lesions (ILs) becomes important to facilitate with radiologists in clinical practice. In this paper, we first study the implementation of the Mask-RCNN model to segment the prostate and ILs. We trained and evaluated models on 120 patients from two different cohorts of patients. We also used 2D U-Net and 3D U-Net as benchmarks to segment the prostate and compared the model's performance. The contour variability of ILs using the algorithm was also benchmarked against the interobserver variability between two different radiation oncologists on 19 patients. Our results indicate that the Mask-RCNN model is able to reach state-of-art performance in the prostate segmentation and outperforms several competitive baselines in ILs segmentation.

## 1. Introduction

Prostate cancer (PCa) is the most common cancer in men in the United Stated (Smith, Andrews et al. 2018). According to the American Cancer Society, 174,650 new cases and 31,620 deaths caused by PCa were expected in 2019 (Siegel, Miller et al. 2019). Currently prostate biopsy is the gold standard to identify the presence of PCa. Due to the poor resolution of gray-scale ultrasound image, conventional transrectal ultrasound (TRUS) guided biopsy has a low detection rates and the rate of false negative can be as high as 40% (Pallwein et al., 2008), which poses a significant threat on patients and is referred to as a "blind" procedure (Ahmed et al., 2009). The advanced technique of MRI-guided biopsy has been introduced and reported by several institutes (Beyersdorff et al., 2005; Anastasiadis et al., 2006; Hambrock et al., 2010; Schoots et al., 2015) for its potential to improve diagnosis accuracy of biopsy and reduce the number of required needle insertion.

Multi-parametic magnetic resonance imaging (mp-MRI), which combines anatomic T1- and T2-weighted imaging (T2WI) with diffusion and perfusion weighted image sequences has been clinically incorporated into prostate cancer diagnosis, risk stratification, staging and treatment guidance. Namely, the apparent diffusion coefficient (ADC) map is derived from diffusion weighted imaging (DWI), which measures the Brownian motion of water molecules in tissue, and depends on many factors such as cell density, size and shape. Perfusion- weighted imaging assesses angiogenesis and microvascular vessel wall permeability. It is dependent on tumor region, temporal and spatial resolution, and the model itself. The use of mp-MRI has been recommended in morphological assessment of the prostate and tumor identification within the gland (Ghai and Haider, 2015).

Automated segmentation of the prostate and screening of prostate cancer from MR images is critical for computer-aided clinical diagnosis, treatment planning and prognosis. However, the development of automatic algorithms remains challenging in several reasons. First of all, there are large variations in image quality caused by several factors at the time of image acquisition (e.g. patient motion, signal-to-noise ratio, use of an endorectal coil, Gadolinium enhancement, etc.). Second, the normal anatomy of the prostate is highly variable across patients and at different time points; and the boundaries between the prostate and surrounding structures (e.g. neurovascular bundles, bladder, rectum, seminal vessels and other soft tissues) are not always immediately clear. The prostate also shows a large variation in size and shape among different patients due to individual differences and physiological changes. Third, the presence of benign conditions such as benign prostatic hyperplasia and prostatitis may mimic the radiographic presentation of a malignancy. Contrast and pixel value of MRI also highlight a large variability in both tissue and texture information.

Over the past few years, progress in image segmentation tasks has been exclusively driven by convolutional neural network (CNN) based models. Many segmentation models fall into two classes. The first class does not rely on the region proposal algorithm. Typical models in this class usually apply an encoder-decoder framework (Liou et al., 2014). The encoder network extracts representations of the image, and the decoder network reconstructs segmentation mask from the learned image representations produced by the encoder network. U-Net (Ronneberger et al., 2015), for instance, is a classic algorithm widely used in biomed- ical image segmentation tasks. Another class of models have their underlying fundamentals on region proposals such as the Mask-RCNN model, which was firstly developed in 2017 by following the basic ideas of Faster-RCNN model. It has a broad use in semantic segmentation, object localization and objection instance segmentation of natural images and human pose estimation (He et al., 2017) and outperformed all existing single-model entries on each task in the 2016 COCO Challenge.

**Technical Significance** We trained and evaluated Mask-RCNN models in the first place to perform multiple tasks including segmentation of both the prostatic gland and ILs. We proposed a method of training for the prostate segmentation to reduce false positives/negatives. Another contribution is that we extended 2D Mask-RCNN model to be able to generate volumetric segmentation of ILs and output top 5 candidate lesions inside the whole prostate.

**Clinical Relevance** Though mp-MRI and other image acquisition tools have being im- proved over decades to capture structural, cellular, vascular and functional properties of prostate, the large amount of data prevents image interpretation, through the prostate de- lineation to ILs localization and segmentation being reproducible and efficient (Lee et al., 2009). Furthermore, subjectivity, variations and fatigue (Dosovitskiy et al., 2015) among interpreters restricts precise and quantitative measurement of ILs prior to therapeutic interventions. In this regard, computer-aided diagnosis systems have been established to help clinicians in their clinical practice. The aim of our research of work is to: 1) provide a quantitative assessment framework for the prostate and ILs; 2) output highly suspicious lesions that is able to augment clinicians to improve accuracy, reduce cost and reduce burnout.

## 2. Patient Cohorts

A total of 120 patients consists of two patient cohorts were studied. **Group A** came from the SPIE-AAPM-NCI Prostate MR Gleason Grade Group Challenge (PROSTATEx-2 Challenge), which is a retrospective set of prostate MR studies conducted by American Association of Physicists in Medicine (AAPM) along with the International Society for Optics and Photonics (SPIE) and the National Cancer Institute (NCI) (Litjens et al., 2014). Each patient case of this database is a single examination from a distinct patient. A random of 78 patients are selected from this database and all patients have biopsy proved lesion status. The images were acquired on two types of Siemens 3T MR scanners, the MAGNETOM Trio and Skyra. Axial

T2WIs were acquired using a turbo spin echo sequence with a resolution of around 0.5 mm in plane and slice thickness of 3.6 mm. Axial DWI sequences were acquired using a single-shot echo planar imaging sequence with a resolution of 2 mm in plane and 3.6 mm slice thickness and with diffusion-encoding gradients in three directions. Three b-values were acquired (50, 400 and 800 s/mm2), and the ADC map was calculated by the scanner software. All images were acquired without an endorectal coil. For group A, the prostate gland and IL based on the biopsy marker were contoured by a clinician from our institute.

**Group B**: Forty-two patients who underwent mp-MRI scans were collected at our institute. An ultrasound guided needle biopsy was performed to confirm the presence of PCa. All examinations were performed using a 3T MR scanner (Ingenia; Philips Medical System, Best, the Netherlands). Axial T2WIs were obtained using Fast-Spin-Echo (TE/TR: 4389/110 ms, Flip Angle: 90◦) with a resolution of 0.42 mm in plane and slice thickness of 2.4 mm. Axial DWIs were obtained (TE/TR: 4000/85 ms, Flip Angle: 90◦) with a resolution 1.79 mm in plane and slice thickness of 0.56 mm. The voxel-wise ADC map was constructed using two DWIs with two b-values (0, 1000 s/mm2). Sixteen patients from group B had prostate gland contoured by a different clinician from group A and all patients had all ILs contoured by three different clinicians from our institute.

In the study of the prostate segmentation, 24 patients were randomly selected from group A and evenly divided into validation set and testing set. The rest of 54 patients were used for the training set. Sixteen patients from group B served as an independent testing cohort. In the study of ILs segmentation, 45 patients from group A and 12 patients from group B were randomly selected into training. Ten patients were selected respectively from group A and group B (for group B we selected patients with only one lesion identified) for validation. The rest 23 patients from group A and 20 patients from group B were used for the testing aim. For the 20 patients from group B, 3 patients have 3 lesions contoured, 8 patients have 2 lesions contoured and the rest 9 patients have only one lesion contoured.

In the study of the prostate segmentation, we used T2WIs only in training, validation and testing since the prostate gland can be well defined on the morphological imaging. The long dimension of T2WI was resized to be 384, and the short dimension was resized to keep the same width/length ratio as the original one and padded by zero to be 384. For ILs segmentation, the combination of T2 and ADC is used as input. The ADC map was re-sampled using bi-linear interpolation and rigidly registered to T2WI using an in-house developed software. The ILs were identified and contoured based on both T2WI and ADC following the criteria of hypointense values on the T2W images and ADC maps. All images were firstly normalized slice by slice, and then had the histogram equalized.

## 3. Methods

### 3.1. Network Architecture

Region proposal network (RPN) is the backbone architecture used as the first stage in both Faster-RCNN and Mask-RCNN and has been proven to be very effective and efficient until now. The purpose of RPN is to identify interested objects within a particular image. One popular backbone architecture is ResNet (He et al., 2016) and its deep variants (e.g. ResNeXt (Xie et al., 2017)) with different depth layers (50 or 101 layers) (Dai et al., 2016; Huang et al., 2017; Liu et al., 2017). Another commonly used backbone architecture is Feature Pyramid Network (FPN) proposed by (Lin et al., 2017). He et al. explored and evaluated different backbones and claimed a ResNet-FPN backbone gained an outstanding performance in both prediction accuracy and time efficiency. Inheriting the spirit of Faster- RCNN, Mask-RCNN also applies bounding box classification and regression in parallel, where the bounding box classifier calculates the probability a proposal containing an object and the regression branch regresses the coordinates of proposals to improve localization accuracy. Mask-RCNN differs from Faster-RCNN that it has an additional branch (head architecture) to output binary mask for each region of interest (RoI). Mask is predicted from each ROI based on the pixel-to-pixel segmentation using a FCN. Mask-RCNN adopts a RoIAlign layer to properly align the extracted features with the input.

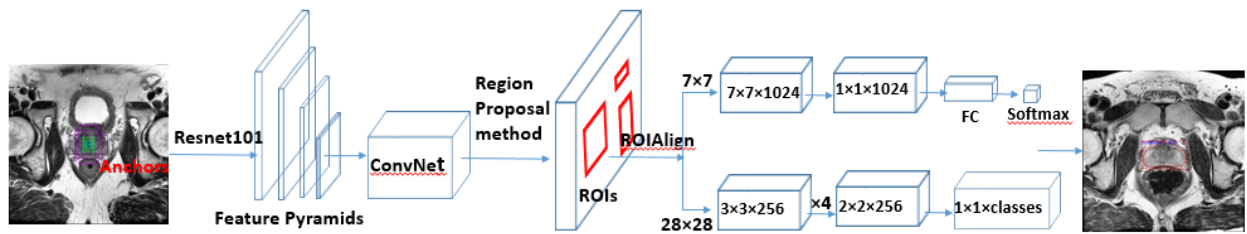

Figure 1: General Mask-RCNN network architecture used in our paper. Predefined an- chors with different scales at one location are shown as purple bounding boxes on the input image. Cubes are represented by kernel size × kernel size × number of filters, above branch is used for classification, bottom branch is used for segmentation.

There are a variety of choices of components for the Mask-RCNN model, and we demonstrate our network architecture in Figure 1. A 101-layers deep residual network is used as the bottom-up pathway for Feature Pyramid Network (FPN) to build pyramid feature maps over the input image and can better extract features at multiple scales. A set of anchors are predefined with relative locations and scales and each one binds to the feature map with corresponding locations and levels. The pyramid features are passed to a shared convolution to generate classes of anchors (background or foreground) and bounding box refinements. Anchors are then filtered by the Region Proposal method, where top anchors are selected, overlapped anchors are removed, and bounding box refinements are again applied to generate final RoIs (which is also called Non-max Suppression). Before running the brunch of RoI classification and bounding box regressor, a ROIAlign algorithm is applied to align feature maps with input properly with a fixed size (e.g. 7 × 7) using interpolation. The bounding box regressor will again refine the location and bounding box size to improve object inclusion. The segmentation branch takes positive RoIs selected by the RoI classifier and generates masks for them. Similarly, feature maps are aligned to input with a fixed size (e.g. 28 × 28). Based on the proximate prostate and PCa size with respect to the image size, anchor sizes are initiated into 16, 32, 64, 128, and 256.

### 3.2. Evaluation Metrics

The metrics we used to evaluate the agreement between the predictions and the given ground truth include the dice similarity coefficient (DSC), 95 percentile Hausdorff distance (HD), sensitivity, and specificity. DSC intends to evaluate how well two binary sets match and a DSC of 1 means the prediction and contoured mask are perfectly matched. HD measures how far two subsets within a metric space are from each other. HD and DSC work as both boundary and volume metrics to provide a more comprehensive view of segmentation accuracy. Sensitivity, also known as the true positive rate (TPR), measures the proportion of actual positives that are correctly identified; specificity, also known as the true negative rate (TNR) measures the proportion of actual negatives that are correctly identified. We also defined the agreement rate to evaluate the degree to which the algorithm segmentation result concur with the segmentation result by clinicians. The formulas for those metrics are shown below:

$$DSC = \frac{2 \cdot \langle y_{true}, \bar{y}_{pred} \rangle}{\langle y_{true}, y_{true} \rangle + \langle \bar{y}_{pred}, \bar{y}_{pred} \rangle},$$

where ytrue is the ground truth contour, y̅pred is the model prediction.

$$HD(A, B) = max_{a \in A}\{min_{b \in B}\{d(a, b)\}\}$$

where a and b are points of sets A and B, respectively, and d(a,b) is Euclidean distance between these points. 95 percentile HD says that 95% of $min_{b \in B}$ d(a,b) is below this amount.

$$Spec. = \frac{DTN}{TN}$$

$$Sens. = \frac{DTP}{TP}$$

where for the prostate segmentation, DTP denotes the number of slices detected the prostate, DTN denotes the number of slices detected no prostate. TP denotes the total number of slices truly containing the prostate, TN denotes the total number of slices truly containing no prostate. For ILs segmentation, DTP denotes the number of detected true positives (detected lesion pixels), DTN denotes the number of detected true negatives (detected background pixels), and TP denotes the total number of positives (total lesion pixels), TN denotes the total number of negatives (total background pixels). Pixels are counted inside the whole prostate.

$$\text{Agreement} = \frac{DTP}{TP}$$

where DTP denotes the number of predicted lesions having DSC greater than 0.2 with lesions contoured by clinicians, TP denotes the number of lesions contoured by clinicians.

### 3.3. Implementation

Our implementation of Mask-RCNN model is based on an existing development by Matter-port Inc., which is based on the open-source libraries Tensorflow and Keras Abdulla (2017). All training, validation and testing phase was conducted on the same workstation equipped with a NVIDIA Quadro P5000 GPU. We trained the network with Adam optimizer Kingma and Ba (2014). The learning rate was initiated to be 0.0001, and β1 equals to 0.9, β2 equals to 0.99. The models were trained for 240,000 iterations; model parameters were saved every 2000 iterations.

As the resolution and field of view (FOV) are different among different MRI scans, several data augmentation options were applied during the training phase, including flipping up to down, flipping left to right, adding random noise, blurring by Gaussian kernel, rotation, horizontal and vertical translation and scaling. With respect to prostate segmentation, flipping left to right was applied. Gaussian noise with a uniform distributed sigma between 0.05 and 0.1 was added to image, and Gaussian kernel with a uniform distributed sigma between 0.8 and 1.3 was applied to blurring image. Rotation angle lied in the range of -20° to 20°. Image was translated at a random value between 0 and 50 pixels in either the vertical direction or horizontal direction. With respect to lesion segmentation, the augmentation methods mentioned above were all applied, but rotation angle was in the range from -80° to 80° and translation was in the range from 0 to 20 pixels. All augmentation options were randomly applied with the probability of 0.5. Models were trained using a batch size of one. DSC was calculated on the validation set which didn't participate in the training phase, and the iteration with the highest DSC was selected as our model. To ensure there is no over fitting and selection bias, we tested the model using the testing set.

### 3.4. Extension of model to 3D space

Many algorithms were developed based on 3D CNN and they claimed it is more effective in exploring volumetric structure. However, we found it difficult to direct extend Mask-RCNN model into 3D architecture as it requires large amount of RPNs and is memory consuming and time consuming to train. We, on the other hand, proposed a method to extend the model to obtain volumetric segmentation results.

**Prostate segmentation** We found 2D Mask-RCNN usually introduces false positive in superior/inferior slices from the prostate gland (bladder and rectum) and false negative in apex and base of the prostate. To train the model for prostate segmentation, an average square area of prostate was estimated based on the training set. For images without a prostate contour, this area was masked and labeled as 'none-prostate'. Images with a prostate contour were masked by the closing contour and labeled as 'prostate'. While in the validation and testing phase, if multiple detections were obtained, we selected the one with the highest score to determine if the current image contained a prostate. We found this way largely decreased the false positive/negative rate.

**Lesion segmentation** We extended the Mask-RCNN model to multi-focal PCa detection and picked top 5 suspicious lesions which is consistent with the procedure of MRI guided prostate biopsy. Several bounding boxes could be detected on the same image for possible localization and segmentation of the ILs. In order to obtain the final detections, we proposed a method to localize the ILs in a recursive manner in the 3D space. Firstly we performed a step similar to non-maxium suppression in object detection, but we calculated the DSC between contours and unified two contours and their bounding box when DSC was greater than 0.5 for each slice. We then selected the top detection with the highest prediction score from all detections within the patient, and then traversed the detection to its adjacent slices whether they were correlated with the top detection. Correlation was determined by the adjacent lesion prediction score and the DSC with the top detection. And two cut off thresholds were fine-tuned on the validation set for the prediction score and DSC. In our experiment, the thresholds of prediction score and DSC were 0.7 and 0.41. The adjacent lesion was then regarded as the top lesion and we recursively traversed all slices to determine a 3D lesion. Finally we eliminated the top 3D lesion from all detections and identified the next top 2 to top 5 lesions following the same steps.

### 4. Results

**Prostate segmentation Results** of prostate segmentation are shown in Table 1. DSC and 95 percentile HD is calculated in the 3D space, while true positives and true negatives are counted slice by slice from all patients. We selected the model that achieved the highest DSC from the validation set and performed testing using both public data and our own data. DSC, 95 HD, TPR and TNR are $0.86 \pm 0.04$, $6.19 \pm 2.38$ (mm), 0.95 and 0.87 respectively using PROSTATEx-2 Challenge dataset, and are $0.82 \pm 0.05$. $8.94 \pm 4.09$ (mm), 0.95 and 0.90, respectively using our own data. Figure 2 shows the prostate segmentation results on T2WI by clinician and the algorithm.

**ILs detection and segmentation** Results of ILs detection and segmentation are shown in Table 2. The validation set has 20 patients/lesions, 10 patients from the PROSTATEx-2 Challenge dataset and 10 patients from our institute with only one lesion identified and contoured. The agreement is 80%, DSC is $0.46 \pm 0.20$. The testing set has 43 patients and 57 lesions, 23 patients from the PROSTATEx-2 Challenge dataset with only one lesion contoured and 20 patients from our institute with multiple lesions contoured. For patients have multiple lesions identified, three patients are identified with three lesions, two of them are detected with two lesions, one of them are detected with one lesion; eight patients have two lesions identified, three of them are detected both. The agreement is 77%, DSC is 0.46 ± 0.20. Figure 3 shows the ILs segmentation results on T2WI by clinician and the algorithm.

**Comparison with U-net** To give a more comprehensive and fair evaluation of the perfor- mance of the Mask-RCNN, we trained a 2D U-Net as well as a 3D U-Net and calculated the DSC on the prostate contours using the same 12 testing patients in prostate segmentation. DSC was calculated to be 0.85±0.03 using 2D U-Net and 0.83±0.07 using 3D U-Net.

Table 1. The result of prostate segmentation by Mask-RCNN

|  | DSC | 95 HD | Sens. | Spec. |
|---|---|---|---|---|
| Validation (12) | 0.88±0.04 | 6.05±2.39 | 0.93 | 0.98 |
| Testing (12) | 0.86±0.04 | 6.19±2.38 | 0.95 | 0.85 |
| Testing with own patients (16) | 0.82±0.05 | 8.94±4.09 | 0.95 | 0.90 |

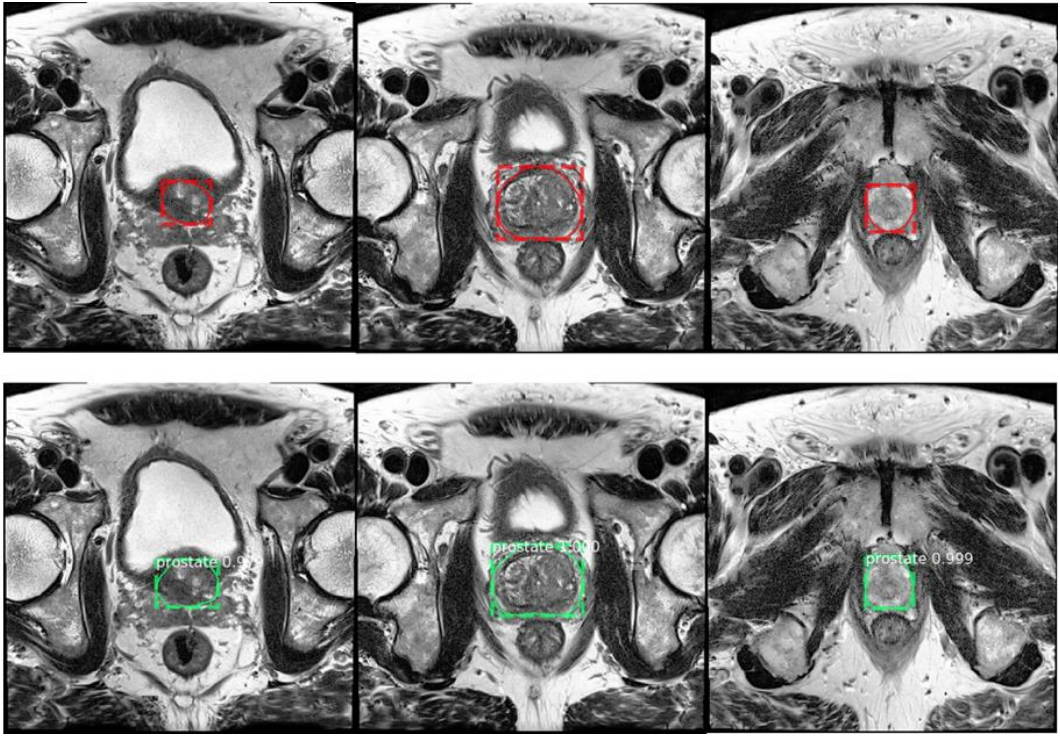

Figure 2: Prostate segmentation result on three slices of T2WI from one patient. Ground truth by the clinician (top rows) is shown with prostate contour and bounding box; Mask- RCNN prediction (bottom rows) is shown with bounding box, prostate contour as well as prediction class and score.

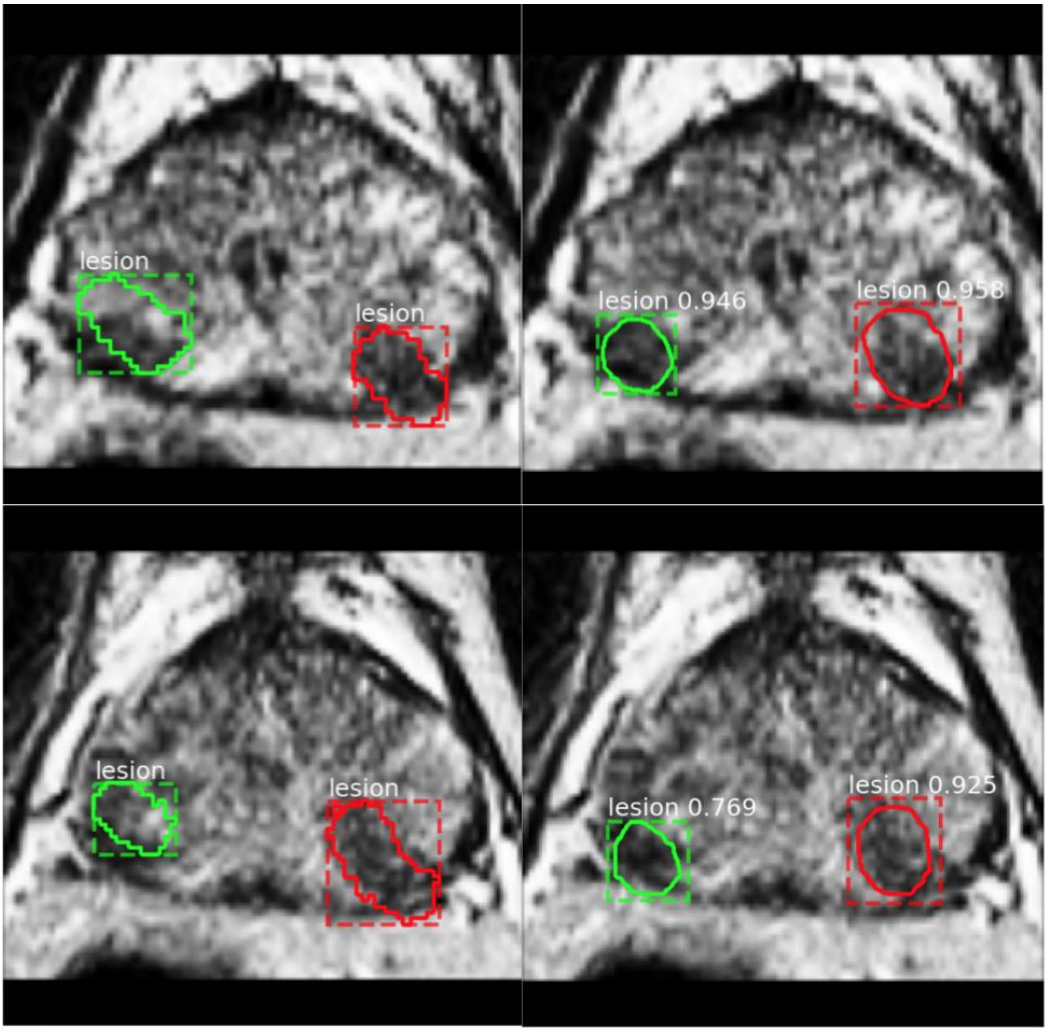

Figure 3: Lesion segmentation on two continuous slices of one patient from our institute with two lesion identified and contoured (green and red) on T2WI. Ground truth by the clinician (left column) is shown with lesion contour and bounding box; prediction of two agreed candidate lesions by the Mask-RCNN (right column) is shown with bounding box, lesion contour as well as prediction class and score.

Table 3. Lesion detection and segmentation results

|  | DSC of detection | Agreement | Sens. | Spec. |
| --- | --- | --- | --- | --- |
| Validation: 10 public patients + 10 own patients (20 lesions) | 0.46 ± 0.20 | 80% | 0.50 ± 0.22 | 0.97 ± 0.01 |
| Testing: 23 public patients + 20 own patients (57 lesions) | 0.44 ± 0.20 | 77% | 0.46 ± 0.22 | 0.97 ± 0.02 |

## 5. Discussion and Related Work

### 5.1. Literature Review

Conventional research applied classic machine learning and statistical graph models for the problem of the prostate and ILs segmentation. The study of neural network based models has only been developed in recent 3 years and still has a long way to go. We performed a literature review to provide a systematic overview over publications and assessed their experiment evaluation set up and results.

| Publication | Task | Method | Result | Evaluation |
|---|---|---|---|---|
| (Tian, Liu et al. 2016) | Prostate segmentation | Graph cut | DSC = 87.0±3.2% | MICCAI 2012 Promise12 challenge |
| (Mahapatra and Buhmann 2014) | Prostate segmentation | Superpixel + Random forests (RF) + Graph cut | DSC = 0.81 | MICCAI 2012 Promise12 challenge |
| (Guo, Gao et al. 2016) | Prostate segmentation | stacked sparse auto-encoder (SSAE) + deformable segmentation | DSC= 0.871±0.042 | 66 T2-weighted MR images |
| (Milletari, Navab et al. 2016) | Prostate segmentation | V-Net + Dice-based loss | DSC = 0.869 ± 0.033 | Trained with 50 MRI volumes Test with 30 MRI scans |
| (Zhu et al., 2017) | Prostate segmentation | *Deeply-Supervised CNN* | DSC = 0.885 | Trained with 77 patients Test with 4 patients |
| (Yu, Yang et al. 2017) | Prostate segmentation | Volumetric convolutional neural network (ConvNet) with mixed residual connections | DSC = 89.43% | MICCAI 2012 Promise12 challenge |
| (Toth and Madabhushi 2012) | Prostate segmentation | landmark-free AAM (MFLAAM) | DSC = 88% ± 5% | Test with 108 studies |
| (Liao, Gao et al. 2013) | Prostate segmentation | Stacked ISA + SL | DSC = 86.7 ± 2.2 | 30 prostate T2 MR images |

| Reference | Task | Method | Results | Dataset |
|---|---|---|---|---|
| (Vincent, Guillard et al. 2012) | Prostate segmentation | Active Appearance Models (AAM) | DSC = 0.88 ± 0.03 | MICCAI 2012 Promise12 challenge |
| (Klein, Van Der Heide et al. 2008) | Prostate segmentation | atlas matching | median DSC varied between 0.85 and 0.88 | leave-one-out test with 50 clinical scans |
| (Li, Li et al. 2013) | Prostate segmentation | random walker | DSC = 80.7±5.1% | 30 dicom volumes of MR images |
| (Kohl, Bonekamp et al. 2017) | ILs Segmentation | Adversarial Networks | DSC = 0.41 ± 0.28<br>Sens. = 0.55 ± 0.36<br>Spec. = 0.98 ± 0.14 | four-fold cross-validation on 55 patients with aggressive tumor lesions |
| (Cameron, Khalvati et al. 2016) | ILs Detection | MAPS (Morphology, Asymmetry, Physiology, and Size) model | ACC = 87%±1%<br>Sens. = 86%±3%<br>Spec. = 88%±1% | 13 patients |
| (Chung, Khalvati et al. 2015) | ILs Segmentation | Radiomics-driven CRF | Sens. = 71.47%<br>Spec.= 91.93%<br>Acc. = 91.17% | 20 patients |
| (Artan, Haider et al. 2010) | ILs Segmentation | Cost-sensitive SVM+CRF | DSC = 39.13%<br>Sens. = 0.84±0.19<br>Spec. = 0.48±0.22 | 21 patients |
| (Artan, Haider et al. 2010) | ILs Localization | random walker | DSC = 0.35±0.18<br>Sens. = 0.51<br>jakkard = 0.44 | 10 patients |
| (Artan, Haider et al. 2010) | ILs Segmentation | RW | Sens. = 0.62±0.23<br>Spec. = 0.89±0.10 | 16 patients with lesions in peripheral zone |
| (Ozer, Haider et al. 2009) | ILs Segmentation | Relevance Vector Machine | DSC = 0.57±0.21<br>Spec. = 0.78<br>Sens. = 0.74 | 20 patients |
| (Artan, Langer et al. 2009) | ILs Segmentation | Cost-sensitive CRF | DSC 0.48<br>Sens. = 0.73±0.25 | 10 patients with lesions only on peripheral zone |

|                          |                      |            | Spec. = 0.75±0.13                                              |             |
|--------------------------|----------------------|------------|----------------------------------------------------------------|-------------|
|                          |                      |            | Acc. = 0.71±0.18                                               |             |
| (Liu, Langer et al. 2009) | ILs Segmentation    | fuzzy MRFs | DSC = 0.45±0.28<br>Spec. = 89.58%<br>Sens. = 87.50%<br>Acc. = 89.38%<br>DSC = 0.6222 | 11 patients |

In this study, we provided a fair evaluation using another patient cohort from our institute to test the algorithm performance. For the prostate segmentation, the validation and testing result shows promising prostate segmentation results using Mask-RCNN on the PROSTATEx-2 Challenge dataset. When using our own data as an independent testing cohort, we observed a slightly decreased performance in DSC of the model as shown in Table 1, while sensitivity and specificity remain equivalent ti the public dataset. We conjecture that it is most likely due to the variability of prostate delineation from a different observer. Other causes may be the variation of image quality acquired on a different scanner from the training dataset as well as small size of dataset. While U-Net is a more elegant fully convolutional network (FCN), it takes the whole image as receptive field for segmentation. Mask-RCNN differs from this kind of segmentation models in that it is based on RPN, which firstly select ROIs and then perform pixel-to-pixel segmentation using FCN on the selected ROI. We provide a more fair comparison between U-Net and Mask-RCNN using the same 12 testing patients from the PROSTATEx-2 Challenge dataset. We compared the DSC calculated by 2D and 3D U-Net with that by the Mask-RCNN, and concluded that Mask-RCNN outperformed slightly both of the 2D and 3D U-Net on prostate segmentation. Mask-RCNN should work better in lesion segmentation than FCN in that it detects possible lesion patches first instead of a direct pixel prediction, which makes it able to output a more similar lesion contour to clinician's contour.

The segmentation of ILs is challenging due to the strong tissue heterogeneities of the prostate and the subtle tumor appearance, leading to large intra- and inter-rater variability in the ground truth. We firstly explored contour differences between two radiation oncologists on same lesions from the same 19 patients and DSC is measured to be $0.67 \pm 0.21$. It proved high inter-observer variability in defining IL boundaries with trained clinicians. Most researchers evaluated their algorithms with less 20 clinical scans, and some focused on lesions in peripheral zone only (Liu et al., 2009; Artan et al., 2009, 2010c), where we validated and tested our algorithm using 63 clinical scans from both public and own patients and segment volumetric lesions inside the whole prostate. As far as we know, the highest performance is achieved by Liu et al. with DSC = 0.6222, Sens. = 87.50%, Spec. = 89.58%. However, the study tested a small cohort of 11 patients and limited to the lesions in peripheral zone only. To our best knowledge, there is only one study applying the neural network based model or ILs segmentation (Kohl et al., 2017) with DSC = $0.41 \pm 0.28$, Sens. = $0.55 \pm 0.36$, Spec. = $0.98 \pm 0.14$, but they performed evaluation only on MRI slices with aggressive PCa only. Similar to (Kohl et al., 2017), we achieved relative low sensitivity and high specificity compared to classic graph models. We assumed there are two reasons: 1) sensitivity and specificity are calculated based on the whole prostate volume instead of a single MRI slice, which leads to the high specificity; 2) classic graphic model predicts lots of false positives, while our algorithm predicts lesions with more reasonable size and shape.

In conclusion, our framework of research is highly consistent with several clinical proce- dures. It is able to work as an end-to-end system to first delineate the prostate gland automatically and prospect for five highly suspicious volumetric lesions inside the whole prostate. The results indicate there is a good agreement between model's selections and clinician's contours, which reveals its capacity in facilitating clinicians in diagnosis, MRI guided biopsy, focal therapy or radiation therapy. It also has huge potential to augment clinicians to improve diagnosis accuracy, reduce labour cost and reduce burnout.